\documentclass[10pt,twocolumn,letterpaper]{article}

\usepackage{cvpr}
\usepackage{times}
\usepackage{epsfig}
\usepackage{graphicx}
\usepackage{amsmath}
\usepackage{amssymb}
\usepackage{algorithm,algorithmicx,algpseudocode}
\usepackage{amssymb}
\usepackage{multirow}
\usepackage{caption}
\usepackage{mathtools}
\usepackage{microtype}
\usepackage{booktabs}

\algnewcommand{\Initialize}[1]{%
  \State \textbf{Initialization:}
  \Statex \hspace*{\algorithmicindent}\parbox[t]{.8\linewidth}{\raggedright #1}
}

\usepackage[pagebackref=true,breaklinks=true,letterpaper=true,colorlinks,bookmarks=false]{hyperref}

\cvprfinalcopy %

\def\cvprPaperID{1027} %

\ifcvprfinal\fi
\begin{document}

\title{Deep Grouping Model for Unified Perceptual Parsing}

\author{Zhiheng Li$^1$ \hspace{1mm} Wenxuan Bao$^2$\thanks{The work was performed while Wenxuan Bao was a visiting student at University of Rochester.} \hspace{1mm} Jiayang Zheng$^1$ \hspace{1mm} Chenliang Xu$^1$\\
$^1$University of Rochester \hspace{1mm} $^2$Tsinghua University\\
\small \{\tt zhiheng.li,jiayang.zheng,chenliang.xu\}@rochester.edu bwx16@mails.tsinghua.edu.cn}

\maketitle

\begin{abstract}
The perceptual-based grouping process produces a hierarchical and compositional image representation that helps both human and machine vision systems recognize heterogeneous visual concepts. Examples can be found in the classical hierarchical superpixel segmentation or image parsing works. However, the grouping process is largely overlooked in modern CNN-based image segmentation networks due to many challenges, including the inherent incompatibility between the grid-shaped CNN feature map and the irregular-shaped perceptual grouping hierarchy. Overcoming these challenges, we propose a deep grouping model (DGM) that tightly marries the two types of representations and defines a bottom-up and a top-down process for feature exchanging. When evaluating the model on the recent Broden$+$ dataset for the unified perceptual parsing task, it achieves state-of-the-art results while having a small computational overhead compared to other contextual-based segmentation models. Furthermore, the DGM has better interpretability compared with modern CNN methods.
\end{abstract}

\section{Introduction}

Deep CNN methods have achieved substantial performance improvement compared with non-CNN methods in the field of semantic segmentation~\cite{Long_2015_CVPR, Chen_2018_PAMI}. Many of them can achieve even better performance by incorporating \textit{good practices} that have long been discovered in non-CNN methods, \eg, multiscale features~\cite{Zhao_2017_CVPR, Xiao_2018_ECCV, Wang_2019_PAMI} and contextual information~\cite{Zhang_2018_CVPR, Yuan_2018_ArXiv, He_2019_CVPR, Fu_2019_ICCV, Zhang_2019_CVPR}. However, recent works still have some key limitations.
First, many CNN-based methods are solely driven by the cross-entropy loss computed against ground-truth pixel labels, lacking an explicit modeling of the perceptual grouping process, which is an integral part in the human visual system~\cite{Brady_2017_JEP}.
Second, most modelings are still focusing on regular-shaped feature maps, which creates not only significant overhead in a multi-scale representation when considering feature-to-feature attention but also is sub-optimal for modeling irregular-shaped semantic regions on the image.

\begin{figure}
    \centering
    \includegraphics[width=1.0\columnwidth]{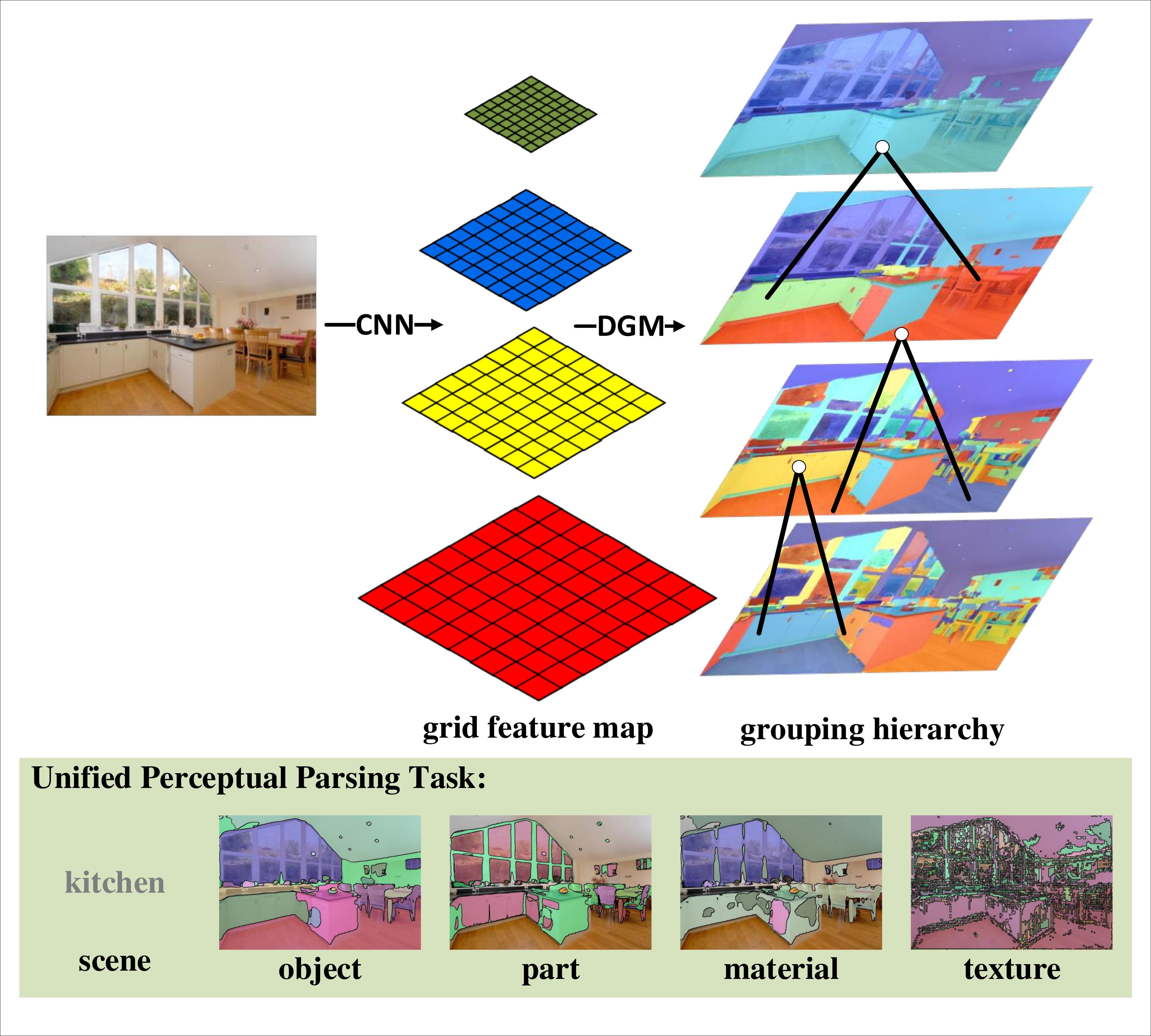}
    \caption{Perceptual grouping process. From fine to coarse: neighboring pixels form a part; parts group into an object; and objects combine into a contextual region. The DGM aims to marry a CNN with the grouping hierarchy for unified perceptual parsing of images. The grouping hierarchy is dynamically computed based on the CNN features, and the CNN features are enhanced by the grouping cues from the graph hierarchy. The model is applied to unified perceptual parsing task to show superiority of DGM.}
    \label{fig:teaser}
    \vspace{-5mm}
\end{figure}

To overcome these limitations, we revisit the classical perceptual grouping methods, \eg, superpixel segmentation~\cite{Ren_2003_CVPR, felzenszwalb2004efficient, moore2008superpixel, Pont-Tuset_2017_PAMI} and image parsing~\cite{tu2005image, tighe2013superparsing, yao2012describing}, which were extensively studied before the predominance of CNNs in segmentation. The seminal work by Tu \etal~\cite{tu2005image} represents an image as a hierarchical graph, a.k.a. \textit{parsing graph}. In their depicted example, an image of \textit{a football match scene} is first decomposed into three elements: person, sports field, and spectator, then these elements are further decomposed, \eg, the person consists of face and body texture. Such a graph is both compositional (\eg, lower-level semantics induce grouping cues for higher-level semantics) and decompositional (\eg, higher-level semantics provide feature support for lower-level semantics), and it varies upon the input image. In this work, we explore whether it is beneficial to inject such a perceptual grouping process explicitly in modern CNN frameworks for a unified image parsing of the scene (see Fig.~\ref{fig:teaser} for an example).

Three challenges arise when incorporating the perceptual grouping process as a hierarchical graph in a deep CNN. First, there is feature incompatibility between the grid-shaped CNN feature maps and irregular-shaped graph nodes, not to mention how to benefit one from the other. Second, it is unclear how to dynamically grow the grouping hierarchy based on different levels of feature semantics extracted from the image. Although superpixel segmentation map provides a plausible initial grouping based on low-level textural and edge cues, high-level semantics of larger receptive fields are needed when growing parts into objects. Third, a holistic understanding of the scene is required when considering the unified pcerceptual parsing task. For example, knowing the scene-level \textit{kitchen} label helps clarify \textit{countertop} against \textit{desk}. It is easy to do in a CNN but difficult in a parsing graph hierarchy.

To tackle the challenges as mentioned above, we propose a novel \textit{Deep Grouping Model (DGM)}, which contains a few modules that are general enough to adapt to many CNNs. The \textit{Expectation-Maximization Graph Pooling} (\textit{EMGP}) module and \textit{Projection} module transform multi-resolution feature maps into a multi-level graph by grouping different regions on the feature map in a bottom-up fashion (\ie, from high- to low-resolution). They have several advantages. Since the model groups pixels and regions iteratively, the number of nodes in the graph is far smaller than the number of pixels on a feature map, which reduces computational overhead. The relationship between different levels of the hierarchy are learned during grouping, rather than assuming a uniform distribution such as in bilinear interpolation or adaptive average pooling on a grid feature map \cite{Xiao_2018_ECCV, Zhao_2017_CVPR}. Furthermore, the contextual information at one level of hierarchy can be quantified via edge weights in a graph, which is sparser than fully-connected non-local block~\cite{Wang_2018_CVPR, Yuan_2018_ArXiv}, leading to a lower overhead.

We put forward a \textit{Top-down Message Passing (TDMP)} module, which propagates contextual information from the top-level graph to the bottom level graph by utilizing grouping results from \textit{EMGP}. In this way, higher level context can be propagated \textit{adaptively} to the corresponding irregular-shaped regions. For instance, object context features (\eg, human) at higher-level graph will be propagated to its corresponding parts (\eg, arms, legs, torso, etc.) at lower-level graph. Similarly, global scene context can also be propagated down to lower-level graph containing objects. Our proposed \textit{TDMP} module is especially useful in the multi-task settings, where lower-level features enhanced by high-level semantics are able to produce better results. At the end, we use \textit{Re-projection} module to re-project features from the hierarchical graph back to multi-resolution grid feature maps, which are used for down-stream tasks.

In order to prove the effectiveness of the proposed model, we apply our model on unified perceptual parsing task, a challenging task to recognize diverse perceptual concepts, including object (or stuff) segmentation, parts segmentation, scene classification, material segmentation, and texture prediction. We use the recent Broden+ dataset~\cite{Bau_2017_CVPR}, a large-scale dataset combining five different datasets with heterogeneous task labels, that is designed for the unified perceptual parsing task. Our method is trained in a multi-task learning fashion, and we evaluate our model on each subtask. Results show that our method achieves the state-of-the-art on Broden+ dataset in every subtask.

Furthermore, the proposed DGM provides better interpretability thanks to the hierarchical graph representation. By using the grouping result , DGM can be applied to other two applications: 1) click propagation, 2) explainability with Grad-CAM, which are the building blocks in recent works on interactive segmentation~\cite{Xu_Ning_2016_CVPR, Majumder_2019_CVPR} and weakly-supervised segmentation~\cite{Wei_2017_CVPR, Wang_Xiang_2018_CVPR, Singh_2017_ICCV}.

\begin{figure*}[t]
  \includegraphics[width=0.85\textwidth]{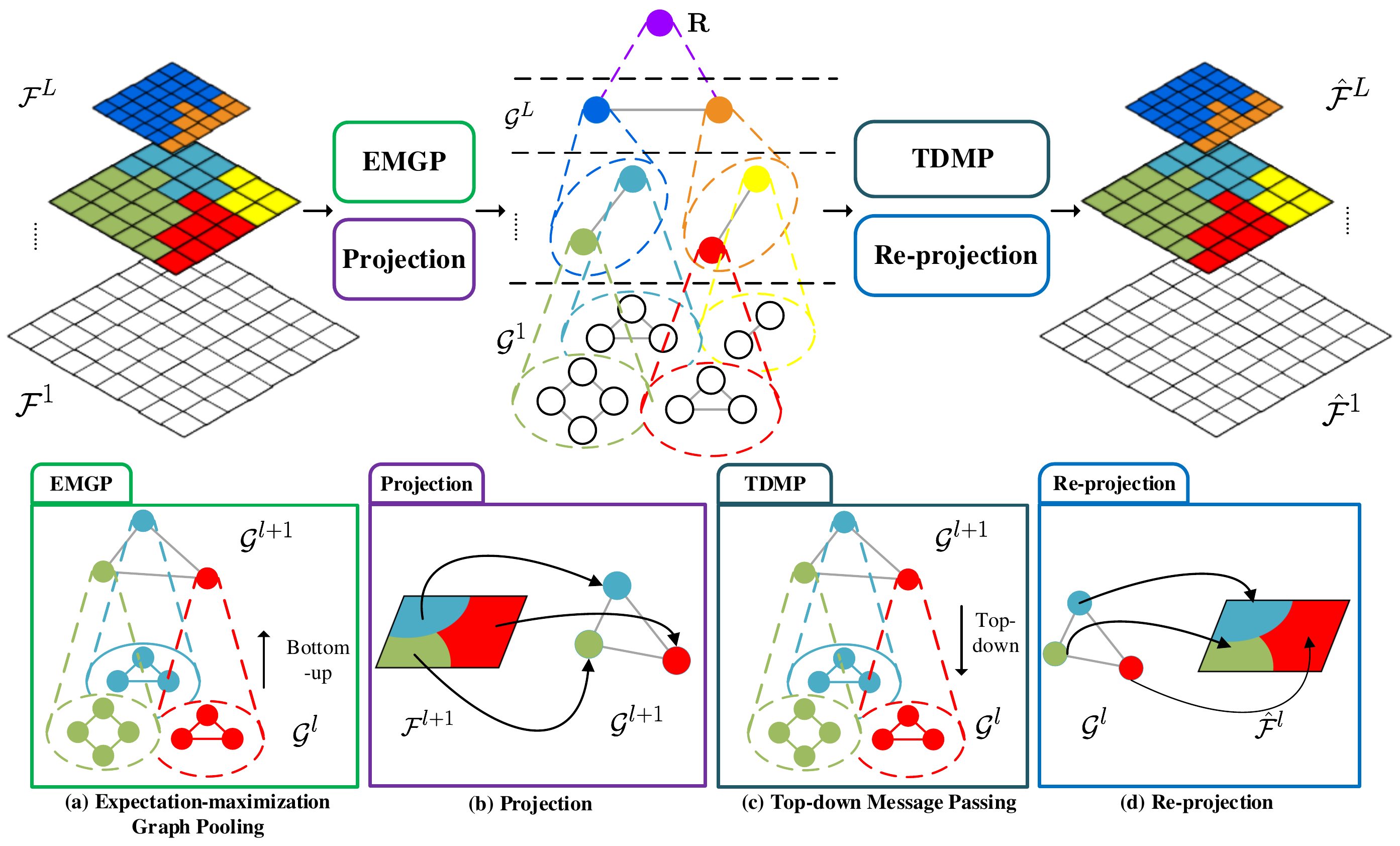}
  \centering
  \caption{An overview of the proposed Deep Grouping Model (DGM).}
  \label{Fig:main_fig}
\end{figure*}

\section{Related Work}
\label{Sec:related_work}

\noindent \textbf{Grouping-based Method.} Grouping-based segmentation method is extensively utilized before the deep learning methods. Ren \etal~\cite{Ren_2003_CVPR} propose grouping pixels into superpixels using Gestalt cues. Hierarchical grouping methods~\cite{Arbelaez_2014_CVPR, Pont-Tuset_2017_PAMI, Uijlings_2013_IJCV, Xu_2016_CVPR, Xu_2013_ICCV, Xu_2012_ECCV} are also proposed for both image segmentation and video segmentation tasks. More recently, some deep learning methods start using grouping in the segmentation task. Gadde \etal~\cite{Gadde_2016_ECCV} use superpixels to upsample CNN's low resolution prediction to the original image size. \cite{Tu_2018_CVPR, Jampani_2018_ECCV} use deep feature rather than traditional low-level cues to predict superpixel map. Two works are closely related to our work. \cite{Hu_2019_ICCV} puts forward local relation layer to model pixel-pair affinity in a predefined $7\times7$ square neighborhood, while our proposed model considers the neighborhood adaptively in an irregular-shaped region. Liang \etal~\cite{Liang_2017_CVPR} propose structure-envolving LSTM where Graph LSTM~\cite{Liang_2016_ECCV} is used for updating node features. In their work, only one pair of nodes is merged each time when a coarser graph is generated. Compared with ~\cite{Liang_2017_CVPR}, our model groups nodes more quickly thus reduces computational overhead. Farabet \etal~\cite{Farabet_2012_PAMI} use multi-scale convolutional feature and conditional random field to regulate the probability of each pixel in segmentation prediction. In contrast, our work learns both grouping hierarchy and top-down message passing at feature level in a end-to-end fashion.

\noindent \textbf{Graph Neural Network.} Some recent works employ Graph Neural Network on segmentation task. Liang \etal~\cite{Liang_2018_NeurIPS} map feature maps to a concept tree to enable concept reasoning. Other works~\cite{Li_2018_NeurIPS, Chen_2019_CVPR} project feature map to graph via linear transformation with learned anchor vectors or convolutional weights, which may be successful in classifying single pixel's semantic meaning but does not consider similarity between pairs of pixels to group them into a region. Ying \etal~\cite{Ying_2018_NeurIPS} propose a differentiable pooling method through predicting pooling weights by GraphSAGE~\cite{Will_2017_NeurIPS}, but the method does not consider pairwise similarity between graph nodes and the number of clusters is also fixed. In comparison, our model considers pairwise affinity among nodes and supports a dynamic number of clustering centers.

\noindent \textbf{Contextual Modeling.} Given the success of self-attention mechanism in many recognition tasks~\cite{Wang_2018_CVPR}, recent work introduces self-attention module in the semantic segmentation field from different perspectives. Yuan \etal~\cite{Yuan_2018_ArXiv} propose object context pooling module. Fu \etal~\cite{Fu_2019_CVPR} apply attention mechanism on both position and channel. The aforementioned non-local based context modeling method creates large overhead since similarity between each pair of grid needs to be computed on the feature map. He \etal~\cite{He_2019_CVPR} introduces adaptive context module to model the affinity between region feature and pixel feature, where the region feature is computed from average pooling on square patch. In comparison with non-local based method and adaptive context module, our method models the context between nodes at different levels of the graph hierarchy, which not only leads to lower overhead but also allow contextual information flow to irregular-shaped regions.

\section{Deep Grouping Model (DGM)}
\label{Sec:DGM}

The proposed DGM represents an image as a hierarchical graph (see Fig.~\ref{Fig:main_fig}). The $L$-level multiscale feature maps $\{\mathcal{F}^l \: | \: l = 1, \dots, L \}$ are extracted from different layers' output of a CNN, where $\mathcal{F}^1$ has a large resolution with more low-level details and $\mathcal{F}^L$ is in the lowest resolution containing more high-level semantics~\cite{Zeiler_2014_ECCV}. Correspondingly, we denote graph feature at the $l$-th level as $\mathcal{G}^l = \langle \mathbf{V}^l, \mathbf{E}^l \rangle$, where $\mathbf{V}^l$ and $\mathbf{E}^l$ denote vertex features and adjacency matrix, respectively. First, we initialize the bottom level graph $\mathcal{G}^1 = \langle \mathbf{V}^1, \mathbf{E}^1 \rangle$ from pre-computed superpixel $\mathcal{S}$ and bottom level grid feature map $\mathcal{F}^1$. Concretely, vertex features come from superpixel pooling, \ie, each node takes the mean of the features in the corresponding superpixel region of the feature map (formal definition can be seen in supplementary material). Unweighted adjacency matrix $\mathbf{E}^1$ is defined from the region adjacency graph of the superpixel $\mathcal{S}$ ~\cite{Tremeau_2000_TIP}, which is much sparser compared with fully-connected non-local operation~\cite{Wang_2018_CVPR, Yuan_2018_ArXiv}. Notice that only $\mathbf{E}^1$ is unweighted adjacency matrix, while upper-level adjacency matrices $\mathbf{E}^l (l > 1)$ are weighted adjacency matrices (more details in Sec.~\ref{Subsec:bottom-up}).

\paragraph{Bottom-up process.}
The bottom-up process is aiming at transforming multi-resolution grid feature maps $\{\mathcal{F}^l \: | \: l = 1, \dots, L \}$ to hierarchical graph representation $\{\mathcal{G}^l \: | \: l = 1, \dots, L \}$ (see Fig.~\ref{Fig:main_fig}), where $\mathcal{G}^l$ not only dynamically composes information from lower level graph $\mathcal{G}^{l-1}$ (Fig.~\ref{Fig:main_fig}(a)), but also receives high-level semantics from feature map $\mathcal{F}^l$ (Fig.~\ref{Fig:main_fig}(b)). To this end, the proposed \textit{Expectation-Maximization Graph Pooling (EMGP)} module and \textit{Projection} module do the aforementioned jobs, respectively.

\paragraph{Top-down process.}
From another perspective, high-level semantics can also help low-level representation. To this end, \textit{Top-down Message Passing} (\textit{TDMP}) module propagates messages from the top-level graph to the bottom-level graph (Fig.~\ref{Fig:main_fig}(c)).

Finally, in order to make DGM compatible with modern CNN framework, we use a \textit{Re-projection} module to transform hierarchical graph $\{\mathcal{G}^l\}$ back to multi-level grid-shape feature map $\{\mathcal{\hat{F}}^l \: | \: l = 1, \dots, L\}$ (Fig.~\ref{Fig:main_fig}(d)), which will be used in down-stream tasks.

\subsection{Bottom-up Graph Hierarchy Construction}
\label{Subsec:bottom-up}

The bottom-up process transforms $\{\mathcal{F}^l\}$ to multi-level graph features $\{\mathcal{G}^l = \big \langle \mathbf{V}^l, \mathbf{E}^l \big \rangle \: | \: l = 1, \dots, L \}$ from the bottom level to the top level (\ie, $l$ is in an increasing order when constructing the graph hierarchy). Concretely, in order to construct $\mathcal{G}^{l+1}$ from $\mathcal{G}^l$, the modules \textit{EMGP} and \textit{Projection} run successively.

\paragraph{Expectation-Maximization Graph Pooling (EMGP).}
The goal of \textit{EMGP} is to pool graph $\mathcal{G}^l$ to $\mathcal{G}^{l+1}$ with less number of nodes, \ie, $|\mathbf{V}^{l+1}| < |\mathbf{V}^{l}|$ (see Fig.~\ref{Fig:main_fig}(a)). Following the EM framework~\cite{Dempster_1977_JRSS}, we initialize $\bar{\mathbf{V}}^{l+1}$ with uniformly sampled vertices from $\mathbf{V}^l$, then update pooled graph vertex features $\bar{\mathbf{V}}^{l+1}$ in $K$ iterations:
\begin{align}
\label{Eq.pool_weight}
&\mathbf{P}^l_{ij} = \frac{1}{\mathbf{Z}^l_j} \exp{(-\frac{{||\mathbf{V}^l_i - \bar{\mathbf{V}}^{l+1}_j ||^2}}{\sigma^2})} \enspace, \\
&\mathbf{\bar{V}}^{l+1} = (\mathbf{P}^l)^\intercal \mathbf{V}^l \enspace,
\end{align}
where $\mathbf{P}^l  \in \mathbb{R}^{|\mathbf{V}^{l}| \times |\bar{\mathbf{V}}^{l+1}| } $ computes the affinity of vertices between the levels $l$ and $l+1$ via a Gaussian kernel with bandwidth $\sigma$ and $\mathbf{Z}^l \in \mathbb{R}^{|\bar{\mathbf{V}}^{l+1}|}$ is a normalization term:
\begin{align}
\mathbf{Z}^l_j = \sum_i^{|\mathbf{V}^l|} \exp{(-\frac{{||\mathbf{V}^l_i - \bar{\mathbf{V}}^{l+1}_j ||^2}}{\sigma^2})}
\enspace.
\end{align}

After $K$-iteration updates of vertex features, following Ying \etal~\cite{Ying_2018_NeurIPS}, the adjacency matrix of higher level graph $\mathbf{E}^{l+1}$ can be computed by:
\begin{align}
\mathbf{E}^{l+1} = (\mathbf{P}^l)^\intercal \mathbf{E}^l \mathbf{P}^l
\enspace.
\end{align}
Notice that our method is different from the ``differentiable pooling'' method proposed in ~\cite{Ying_2018_NeurIPS}. Instead of predicting pooling weights $\mathbf{P}^l$ through a stack of graph convolutional layers, our method uses EM to make the prediction. Therefore, our method not only considers similarity between each pair of nodes, but also can change $|\mathbf{V}^{l+1}|$ dynamically according to the content of the image. For example, an image of a simple scene with small number of objects or uniform textual, \eg, the \textit{sky}, can be represented by a small $|\mathbf{V}^{l+1}|$ in the graph.

\paragraph{Projection.} Although the pooled node features $\bar{\mathbf{V}}^{l+1}$ summarize the lower level graph through a linear combination of the lower level graph nodes $\mathbf{V}^l$, they do not necessarily contain higher level semantics. To incorporate higher level semantics, the \textit{Projection} module projects the feature map $\mathcal{F}^{l+1}$ to pooled node features $\bar{\mathbf{V}}^{l+1}$, outputting node feature $\mathbf{V}^{l+1}$.

A straightforward design could be constructing a bipartite graph between $\mathcal{F}^{l+1}$ and $\bar{\mathbf{V}}^{l+1}$ and use graph convolution to propagate high-level semantics, where pixels on the feature map $\mathcal{F}^{l+1}$ are treated as nodes and directed edges are pointing from $\mathcal{F}^{l+1}$ to $\bar{\mathbf{V}}^{l+1}$. However, such design not only creates large overhead due to large number of pixels on the feature map, but also the edge weights of the bipartite graph is undefined. Therefore, we define auxiliary nodes $\mathbf{U}^{l+1}$, obtained from superpixel pooling on feature map $\mathcal{F}^{l+1}$ by the bottom-level superpixel map $\mathcal{S}$, to address the aforementioned problems. Since both $\mathbf{U}^{l+1}$ and $\mathbf{V}^1$ are computed from the same superpixel map $\mathcal{S}$, $\mathbf{U}^{l+1}$ has the same number of vertices as $\mathbf{V}^1$, \ie, $|\mathbf{U}^{l+1}| = |\mathbf{V}^1|$. However, $\mathbf{U}^{l+1}$ contains high-level semantics as it is pooled from the feature map $\mathcal{F}^{l+1}$.

A quasi-bipartite graph from $\mathbf{U}^{l+1}$ to $ \bar{\mathbf{V}}^{l+1}$ can be constructed. Since $\mathbf{U}^{l+1}$ can also be hierarchically grouped to $\mathbf{V}^{l+1}$ as how $\mathbf{V}^1$ are merged to $\mathbf{V}^{l+1}$, we reuse $\{ \mathbf{P}^l \}$ predicted by \textit{EMGP} to construct the adjacency matrix of the quasi-bipartite directed graph. Concretely, we compute the cumulative product
$\prod_{k=1}^l \mathbf{P}^k \in \mathbb{R}^{ |\mathbf{V}^1| \times |\mathbf{V}^{l+1}| }$, which can be regarded as graph pooling weights that directly pool $\mathbf{V}^1$ (or the auxiliary nodes $\mathbf{U}^{l+1}$) to $\mathbf{V}^{l+1}$. To enable vertices $\bar{\mathbf{V}}^{l+1}$ retain the information through \textit{EMGP}, self-loops are added to $\bar{\mathbf{V}}^{l+1}$, resulting in the final adjacency matrix $\mathbf{I} + \prod_{k=1}^l \mathbf{P}^k$ of the bipartite graph. Therefore, the bipartite graph is formally defined as $ \big( \mathbf{U}^{l+1}, \bar{\mathbf{V}}^{l+1}, \mathbf{I} + \prod_{k=1}^l \mathbf{P}^k \big)$, where directed edges are pointing from $\mathbf{U}^{l+1}$ to $\bar{\mathbf{V}}^{l+1}$.

Next, we use graph convolution to allow message passing from $\mathbf{U}^{l+1}$ to $\mathbf{V}^{l+1}$:
\begin{align}
\label{Eq:projection}
\mathbf{V}^{l+1} = \textit{GConv}\big( \mathbf{U}^{l+1} \cup \bar{\mathbf{V}}^{l+1}, \mathbf{I} + \prod_{k=1}^l \mathbf{P}^k \big)
\enspace,
\end{align}
where \textit{GConv} stands for graph convolution. Following the mean aggregator proposed in GraphSAGE~\cite{Will_2017_NeurIPS}, we use weighted average aggregator GraphSAGE as the graph convolution layer:
\begin{align}
\label{Eq:graphsage}
  \mathbf{h}_v = \sigma(\mathbf{W} \cdot \sum_{u \in \mathcal{N}(v)} \mathbf{w}(u, v) \cdot \mathbf{h}_u)
  \enspace,
\end{align}
where $\mathbf{h}_v$ stands for the feature of vertex $v$, $\sigma$ is the sigmoid function, $\mathbf{W}$ is a learnable weight matrix, and $\mathcal{N}(v)$ defines the neighboring nodes of vertex $v$. Here, $\mathbf{w}(u,v)$ is the weight of the directed edge from $u$ to $v$, which can be found in the given adjacency matrix (\ie, $\mathbf{I} + \prod_{k=1}^l \mathbf{P}^k$ as in Eq.~\ref{Eq:projection}). Thus, the updated graph node features $\mathbf{V}^{l+1}$ contain features of both high-level semantics $\mathcal{F}^{l+1}$ and the feature summarization from its lower level graph $\mathcal{G}^l$.

\paragraph{Global Vector.} After the construction of $\mathcal{G}^L$, we obtain the global vector representation $\mathbf{R}$ (see the top node in Fig.~\ref{Fig:main_fig}) of the scene by:
\begin{align}
\label{Eq:readout}
\mathbf{R} = \textit{READOUT}\big ( \mathbf{V}^L \big)
\enspace,
\end{align}
where \textit{READOUT} function is used for combining features of a graph in many GNN methods~\cite{Ying_2018_NeurIPS, Velickovic_2019_ICLR}. Here we use average pooling as the \textit{READOUT} function. In other words, $\mathbf{R} = \frac{1}{|\mathbf{V}^L|} \sum_i^{|\mathbf{V}^L|}\mathbf{V}^L_i$. $\mathbf{R}$ can also be regarded as a graph at level $L+1$ without edges, \ie, $\mathbf{R} = \mathcal{G}^{L+1} = \langle \mathbf{V}^{L+1}, \emptyset \rangle$. Since $\mathbf{R}$ is a vector representation of the image, it can be supervised by image classification tasks, \eg, a scene category label for the image.

\subsection{Top-down Message Passing (TDMP)}

To further enable high-level semantics to help low-level features, the \textit{TDMP} module iteratively updates each level of graph features from the top-level graph $\mathbf{R} = \mathcal{G}^{L+1}$ to the bottom level graph $\mathcal{G}^{1}$ through message passing, outputting updated multi-level graph features. It serves much like the ``decomposition'' process as motivated in Introduction.

Concretely, given $\mathbf{V}^{l+1}$ (already updated) and $\mathbf{V}^l$ (to be updated), a quasi-bipartite graph is constructed (see Fig.~\ref{Fig:main_fig}(c)), where directed edges are pointing from $\mathbf{V}^{l+1}$ to $\mathbf{V}^l$. Intuitively, high-level semantics should be transmitted to their corresponding lower-level regions. For example, the whole human body feature at the $(l+1)$-th level should be sent to human parts (\eg, arms, legs) at the $l$-th level . Thus, by reusing the grouping results in the bottom-up process, edges $\tilde{\mathbf{P}}^l \in \mathbb{R}^{|\mathbf{V}^{l}| \times |\mathbf{V}^{l+1}|}$ can be obtained by:
\begin{equation}
\tilde{\mathbf{P}}^l_{ij} = \frac{1}{\tilde{\textbf{Z}}^l_i} \exp{(-\frac{{||\mathbf{V}^{l}_i - \mathbf{V}^{l+1}_j ||^2}}{\sigma^2})} \enspace,
\end{equation}
where $\tilde{\textbf{Z}}^l \in \mathbb{R}^{|\bar{\mathbf{V}}^{l}|}$ is a normalization term:
\begin{equation}
    \tilde{\textbf{Z}}^l_i = \sum_j^{|\mathbf{V}^{l+1}|} \exp{(-\frac{{||\mathbf{V}^{l}_i - \mathbf{V}^{l+1}_j ||^2}}{\sigma^2})}
    \enspace.
\end{equation}

After adding self-loops to $\mathbf{V}^l$, a graph convolution layer is applied to achieve the top-down message passing:
\begin{equation}
\mathbf{V}^l := \textit{GConv}\big( \mathbf{V}^{l+1} \cup \mathbf{V}^l , \mathbf{I} + \tilde{\mathbf{P}}^l \big) \enspace,
\end{equation}
where $\mathbf{V}^l$ is the updated vertex feature at the $l$th level and \textit{GConv} is defined the same as in Eq.~\ref{Eq:graphsage}.

\subsection{Re-projection from Graph to Grid Features}

Finally, we re-project the updated vertex features $\{\mathbf{V}^l\}$ back to the grid features resulting in $\{\hat{\mathcal{F}}^l\}$. The \textit{re-prejection} can be regarded as a mirror module of \textit{projection}. Analogous to the \textit{projection} module, at each level $l$, a quasi-bipartite directed graph $\big( \mathbf{V}^l,  \mathbf{U}^l, \mathbf{I} + \prod_{k=1}^l \tilde{\mathbf{P}}^k \big)$ (see Fig.~\ref{Fig:main_fig}(d)) is built from superpixel pooling features $\mathbf{U}^l$, updated vertex features $\mathbf{V}^l$, and the adjacency matrix that comes from the self-loops of $\mathbf{U}^l$ and the cumulative product $\prod_{k=1}^l \tilde{\mathbf{P}}^k$. Here, edges are pointing from $\mathbf{V}^l$ to $\mathbf{U}^l$. Then, we apply graph convolution to re-project the features:
\begin{equation}
\hat{\mathbf{U}}^l = \textit{GConv}\big( \mathbf{V}^l \cup \mathbf{U}^l, \mathbf{I} + \prod_{k=1}^l \tilde{\mathbf{P}}^k \big)
\enspace,
\end{equation}
where $\hat{\mathbf{U}}^l$ is the vertex feature receiving information from the graph and has the same number of superpixels defined in superpixel map $\mathcal{S}$. Lastly, $\hat{\mathbf{U}}^l$ is copied to pixel regions defined by the superpixel map $\mathcal{S}$, outputting the updated grid feature map $\hat{\mathcal{F}}^l$.

\section{Unified Perceptual Parsing with DGM}
\label{Sec:upp_full_model}

\begin{figure}[t]
  \includegraphics[width=8cm]{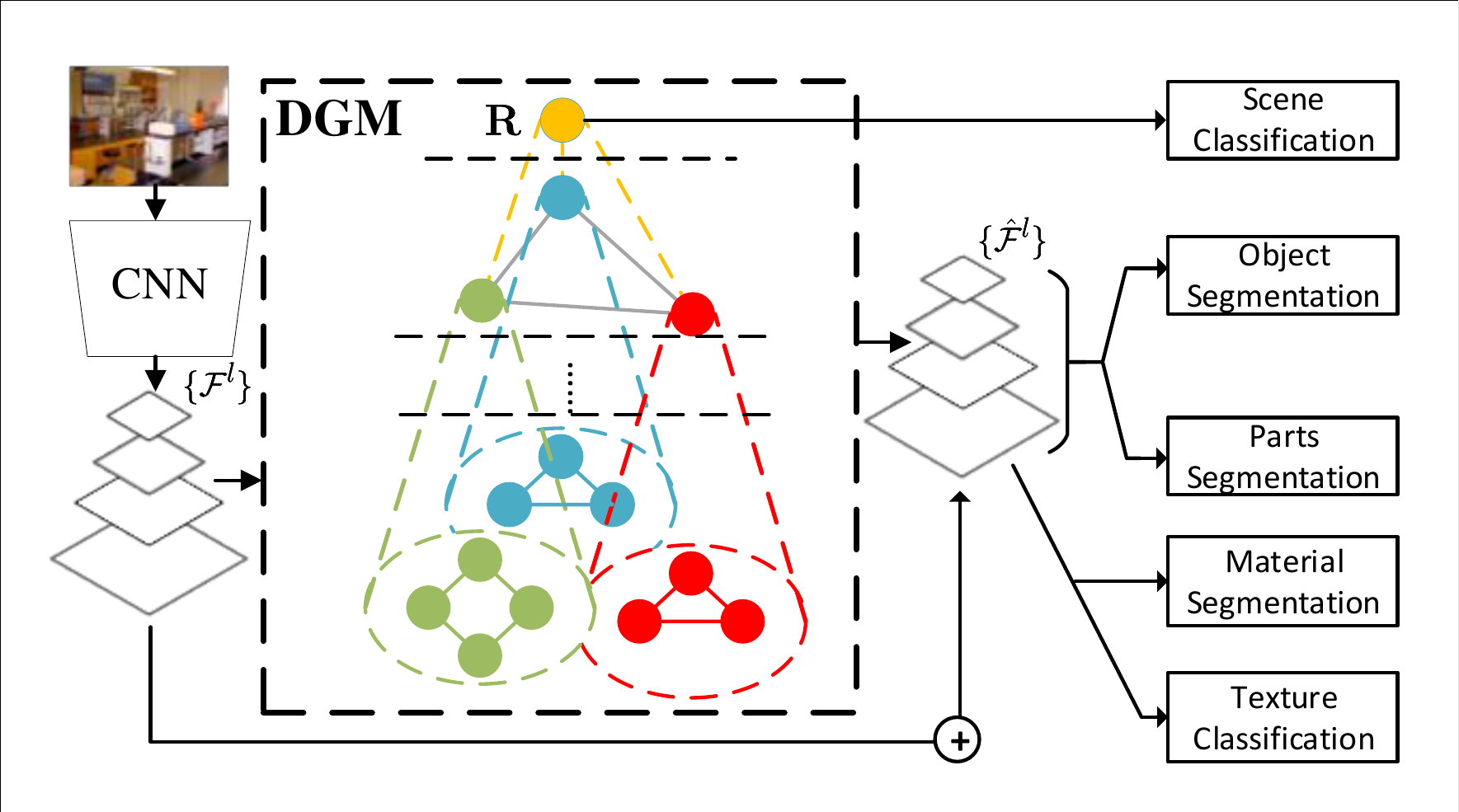}
  \centering
  \caption{Full Model for the Unified Perceptual Parsing Task.}
  \label{Fig:task}
\end{figure}

To fully verify the effectiveness of the hierarchical graph representation, we apply deep grouping model (DGM) on the unified perceptual parsing (UPP) task, a challenging task introduced by Xiao \etal~\cite{Xiao_2018_ECCV}. Aiming at recognizing heterogeneous perceptual concepts of an image, UPP combines tasks of scene classification, object segmentation, parts segmentation, material segmentation, and texture recognition, requiring good modeling on features at different granularities.

To this end, we insert DGM to a backbone model (see Fig.~\ref{Fig:task}), which outputs $\{\mathcal{\hat{F}}^l \: | \: l = 1, \dots, L\}$. With the residual connection~\cite{He_2016_CVPR} from $\{\mathcal{F}^l \: | \: l = 1 ... L\}$, we obtain multi-resolution grid feature maps $\{\mathcal{F}^l + \mathcal{\hat{F}}^l \: | \: l = 1, \dots, L\}$. Following the architecture proposed by Xiao \etal~\cite{Xiao_2018_ECCV}, after bilinear interpolating all feature maps to the same size, we concatenate all $L$ levels of grid features $\{\mathcal{F}^l + \mathcal{\hat{F}}^l \: | \: l = 1, \dots, L\}$ for object segmentation and part segmentation. In material segmentation, we only use the bottom-level grid feature $\mathcal{F}^1 + \mathcal{\hat{F}}^1$ for prediction by following the architecture of UPerNet~\cite{Xiao_2018_ECCV}.

For scene classification, we first apply global average pooling on the original top-level feature map $\mathcal{F}^L$ (not shown in the figure). Then, it is residual connected with the graph \textit{READOUT} feature $\mathbf{R}$ for scene classification.

Limited by the dataset in UPP task~\cite{Bau_2017_CVPR}, only texture images with image-level labels are provided. Therefore, for texture recognition, the model classifies texture images with the feature come from global average pooling on the bottom grid features $\mathcal{F}^1 + \hat{\mathcal{F}}^1$ in training and quantitative evaluation. However, we can also apply the texture classification layer on each pixel to generate texture segmentation results on natural images.

To summarize, the final loss of the full model on the unified perceptual parsing task is defined by:
\begin{equation}
  \mathcal{L} = \lambda_s \mathcal{L}_{s} + \lambda_t \mathcal{L}_{t} + \lambda_o \mathcal{L}_{o} + \lambda_p \mathcal{L}_{p} + \lambda_m \mathcal{L}_{m} ,
\end{equation}
where $\mathcal{L}_{s}$ and $\mathcal{L}_{t}$ are cross-entropy losses between prediction and image labels for scene classification and texture classification, respectively. $\mathcal{L}_{o}$, $\mathcal{L}_{p}$, $\mathcal{L}_{m}$ are cross-entropy losses at each pixel between the prediction and ground-truth for object segmentation, part segmentation, and material segmentation, respectively. Following~\cite{Xiao_2018_ECCV}, coefficients of each loss term are $\lambda_s=0.25, \lambda_t=1, \lambda_o=1, \lambda_p=0.5, \lambda_m=1$.

\begin{table*}[t]
  \centering
  \begin{tabular}{l|c|cc|cc|c|cc|c}
  \toprule
  \multirow{2}{*}{Tasks}     & \multirow{2}{*}{Method}                                & \multicolumn{2}{c|}{Object} & \multicolumn{2}{c|}{Part} & Scene      & \multicolumn{2}{c|}{Material} & Texture    \\ \cline{3-10}
                             &                                                               & mIoU         & P.A.         & mIoU        & P.A.        & Top-1 Acc. & mIoU          & P.A.          & Top-1 Acc. \\ \midrule
  \multirow{5}{*}{O+P+S}     & APCNet                                                        & 21.25        & 71.71        & 23.39       & 41.07       & 68.50      & -             & -             & -          \\
                             & OCNet                                                         & 22.62        & 74.58        & 28.51       & 48.92       & 68.50      & -             & -             & -          \\
                             & UPerNet                                                       & 23.83        & \textbf{77.23}        & 30.10       & 48.34       & 71.35      & -             & -             & -          \\
                             & \textbf{DGM w/o \textbf{$\downarrow$}}                       & 24.58        & 74.76        & 31.23       & \textbf{51.17}       & 71.24      & -             & -             & -          \\
                             & \textbf{DGM}                       & \textbf{24.76}        & 75.15        & \textbf{31.26}       & 50.55       & \textbf{71.87}      & -             & -             & -          \\ \hline
  \multirow{5}{*}{O+P+S+M+T} &  APCNet                                                        & 20.37        & 71.01        & 22.32       & 40.08       & 68.45      & 43.88             & 79.95             & 50.35 \\
  & OCNet                                             & 20.21        & \textbf{77.09}        & 25.75       & 43.78       & 66.92      & 48.20         & 80.70         & 51.95 \\
   & UPerNet                                          & 23.36        & \textbf{77.09}        & 28.75       & 46.92       & 70.87      & 54.19         & 84.45         & 57.44$^*$      \\
                             & \textbf{DGM w/o $\downarrow$} & 24.05        & 74.21        & 29.94       & 49.49       & 70.24      & 54.52         & 84.41         & 58.15      \\
  & \textbf{DGM}  & \textbf{24.37}        & 74.99        & \textbf{30.28}       & \textbf{49.70}       & \textbf{71.03}      & \textbf{54.58}         & \textbf{84.62}         & \textbf{60.10}          \\ \bottomrule
  \end{tabular}
  \caption{Comparing with state-of-the-art methods on Broden+ dataset. O+P+S means object segmentation task, part segmentation task, and scene classification task are used in training and evaluation. O+P+S+M+T incrementally add material segmentation task and texture classification task in training and evaluation stages. $^*$Based on the authors' released model, we continue to train UPerNet and get better results on texture classification than the reported number (35.10) in ~\cite{Xiao_2018_ECCV}.}
  \label{Tab:Broden}
  \vspace{-4mm}
\end{table*}

\begin{table}[t]
\centering
\begin{tabular}{c|c|c}
\toprule
Method  & mIoU  & Pixel Accuracy \\ \midrule
UPerNet                    & 42.66          & 81.01          \\
\textbf{+DGM w/o $\downarrow$}          & \textbf{43.64} & 81.11 \\
\textbf{+DGM}           & 43.51          & \textbf{81.13}          \\ \hline
HRNetv2                    & 43.20          & 81.47         \\
\textbf{+DGM w/o $\downarrow$}           & \textbf{43.86}          & \textbf{81.55}   \\
\textbf{+DGM}           & 43.46          & 81.53         \\\hline
DeepLabV3                 & 44.1          & 81.1 \\
\textbf{+DGM w/o $\downarrow$}               & 44.31          & \textbf{81.36}  \\
\textbf{+DGM}               & \textbf{44.86}          & 81.35  \\\hline\hline

CCNet~\cite{Huang_2019_ICCV} & 45.22          & -  \\
APCNet~\cite{He_2019_CVPR}               & 45.38          & -  \\
OCNet~\cite{Yuan_2018_ArXiv}            &  45.45  &  - \\ \bottomrule

\end{tabular}
\caption{Results on ADE20k validation set.}
\label{Tab:ade20k}
\vspace{-6mm}
\end{table}

\section{Experiments}
\subsection{Dataset and Evaluation Metrics}
The Broden+ dataset~\cite{Bau_2017_CVPR} is used for training and evaluation on the unified perceptual parsing task~\cite{Xiao_2018_ECCV}. The dataset is comprised of five large datasets: ADE20k~\cite{Zhou_2019_IJCV}, PASCAL Context~\cite{Mottaghi_2014_CVPR}, PASCAL-Part~\cite{Chen_2014_CVPR}, OpenSurfaces~\cite{Bell_2013_SIGGRAPH}, and DTD~\cite{Cimpoi_2014_CVPR} datasets. For each subtask in unified perceptual parsing, the data source comes from the union of the datasets that contain subtask's labels. For example, object/stuff segmentation task will be trained on and evaluated on the union of ADE20k and PASCAL-Context datasets. In this way, not only the number of tasks is large, but also the number of categories is larger since datasets are merged together, which makes the unified perceptual parsing task extremely challenging. In terms of evaluation metrics, the scene classification task and texture classification task are evaluated via top-1 accuracy (Top-1 Acc.). The object/stuff segmentation, parts segmentation and material segmentation are evaluated by mIoU and pixel accuracy (P.A.).

\begin{figure}[t]
  \includegraphics[width=\columnwidth]{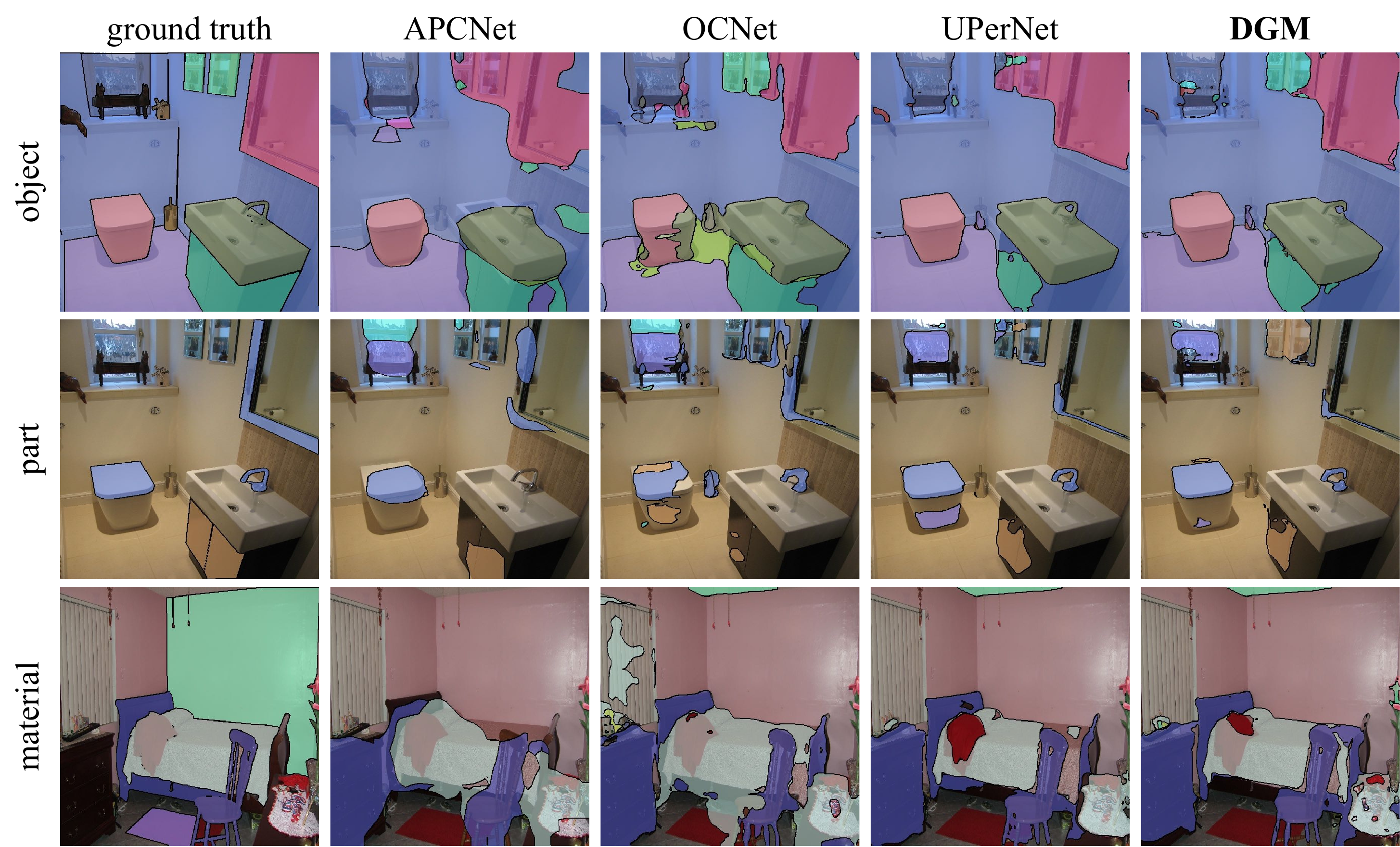}
  \centering
  \caption{Qualitative comparison on Broden+ Dataset.}
\label{Fig:qualitative_broden}
\vspace{-5mm}
\end{figure}

\label{Sec:experiment}
\subsection{Implementation Details}

We follow the experimental settings in~\cite{Xiao_2018_ECCV}. During training, we resize the image's shorter side to a size that is randomly chosen from ${300, 375, 450, 525, 600}$ and keep its aspect ratio. The shorter side of the image is resized to 450 pixels in the evaluation stage. Following~\cite{Chen_2018_PAMI}, we use ``poly'' learning rate policy ($1 - \frac{iter}{max\_iter}^{power}$) to adjust learning rate during training, and the initial learning rate is 0.02, where $max\_iter = 2 \times 10^5$ and $power = 0.9$. The batch size is 8, and the model is trained on 4 GPUs.

MCG~\cite{Pont-Tuset_2017_PAMI, Arbelaez_2014_CVPR} is used for extracting superpixels for training DGM, which is further merged greedily to make sure that the number of superpixel is at most 512.
In terms of the DGM architecture, the input multi-resolution feature map $\mathcal{F}^l$ comes from $C_1$ to $C_4$ layers' output from ResNet~\cite{He_2016_CVPR}. Accordingly, we set the level of graph $L = 4$ in our experiment. All GraphSAGE~\cite{Will_2017_NeurIPS} layers in DGM are followed by $L_2$ normalization and ReLU~\cite{Nair_2010_ICML}. The \textit{EMGP} module pools the graph to half the number of nodes in upper-level graph (\ie, $|\bar{\mathbf{V}}^{l+1}| = |\mathbf{V}^l|/2$). The number of iteration $K$ in \textit{EMGP} is set as 5 in training and 10 in evaluation.

Our code is based on the PyTorch framework~\cite{Paszke_2017_ICLRW}. Specifically, the PyTorch Geometric~\cite{Matthias_2019_ICLRW} is used to implement graph operations in DGM. Following UPerNet~\cite{Xiao_2018_ECCV}, in the experiment on Broden+ dataset, all tasks except the texture classification task are trained jointly. When training the texture classification task, the model's parameters are fixed except the texture classification branch.

\subsection{Comparison with the State-of-the-art}
\label{Sub:}
Results of all tasks in the unified perceptual parsing (UPP) task are shown in Tab.~\ref{Tab:Broden}. Since the dataset is fairly recent and only UPerNet~\cite{Xiao_2018_ECCV} reports its results, we replicate OCNet~\cite{Yuan_2018_ArXiv} and APCNet~\cite{He_2019_CVPR}'s results\footnote{To ensure a fair comparison, we used the authors' released code for OCNet. Since no released code for APCNet, we did our best to replicate.} on the Broden+ dataset, as they represent state-of-the-art contextual modeling methods based on non-local block and region-based context modeling in semantic segmentation, respectively. The backbone of UPerNet, OCNet and our proposed DGM is ResNet50, and APCNet's backbone is the dilated ResNet50~\cite{Xiao_2018_ECCV, He_2019_CVPR, Yuan_2018_ArXiv}. Backbones' weights are initialized with ImageNet~\cite{Deng_2009_CVPR} pretrained models. More results comparing with GCU~\cite{Li_2018_NeurIPS} and HRNetv2~\cite{Wang_2019_PAMI} backbone are included in Appendix.

Results shows that our model (\textbf{DGM} in Tab.~\ref{Tab:Broden}) outperforms all other methods, achieving the state-of-the-art result on Broden+ in every subtask. Although DGM did not achieve the best performance in terms of pixel accuracy on the object segmentation subtask, we suspect that the pixel accuracy measure is easily biased by the imbalanced number of pixels among different classes, while mIoU is a better and more meaningful evaluation metric for segmentation.

In the qualitative evaluation, our model can achieve more reasonable results. For example, in Fig.~\ref{Fig:qualitative_broden}, compared with other methods, our model successfully segments both cabinet (in green) and toilet (in pink) in object segmentation. Our model's parts segmentation has smaller false prediction on the toilet. Finally, on the material segmentation, our model shows sharp boundary on the legs of wood chair.

\begin{figure}[t]
  \includegraphics[width=\columnwidth]{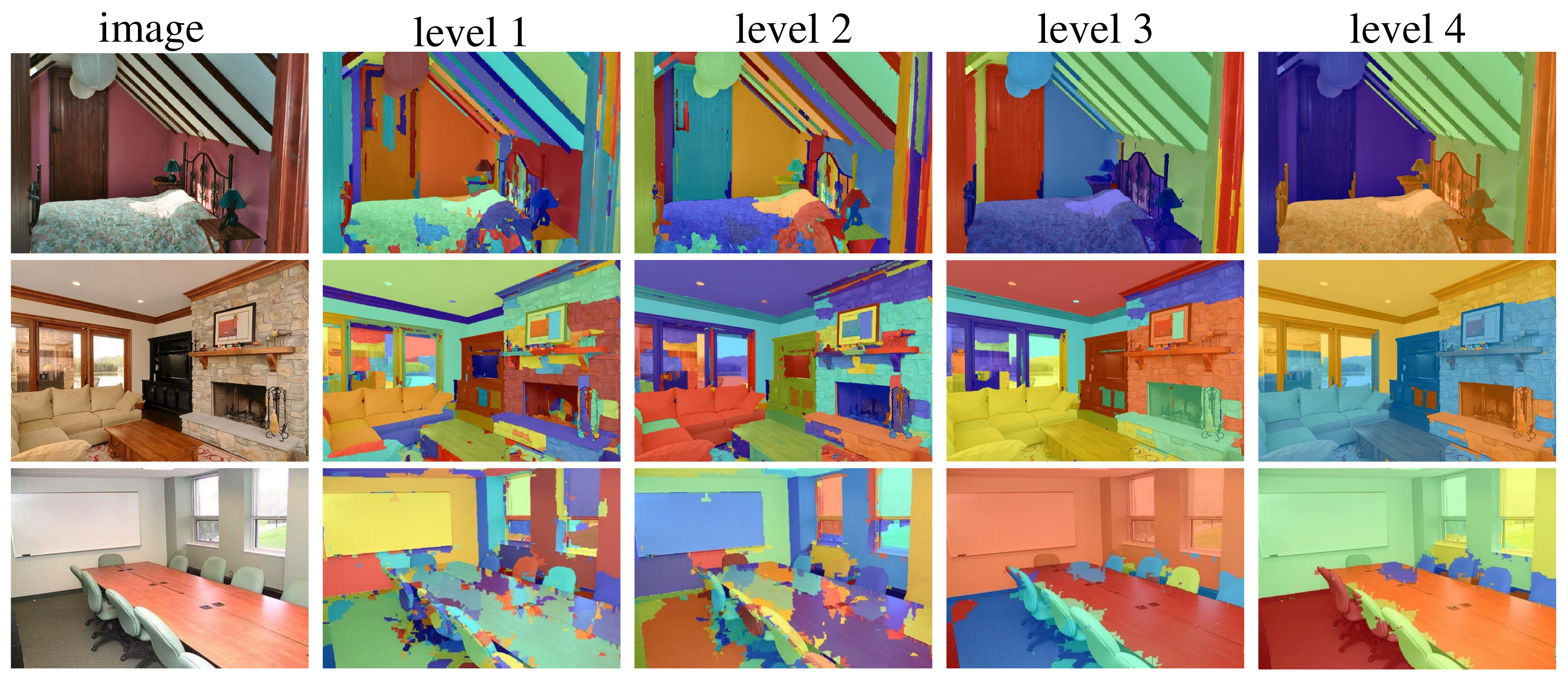}
  \centering
  \caption{Visualization of perceptual groupings generated by DGM. A color represents a graph vertex. Note that the same color between different levels are not related.}
  \label{Fig:group_visualization}
  \vspace{-4mm}
\end{figure}

\subsection{Ablation Study}
\label{Sec:ablation}
\paragraph{Single-task training.}
To ablate the effect of multi-task training, we train our model on ADE20k only focusing on the semantic segmentation task. We use three backbone models to train and evaluate our model: UPerNet, HRNetv2~\cite{Wang_2019_PAMI}, and DeepLabV3~\cite{Chen_2017_ArXiv}. Our DGM is general enough to be an add-on module for many segmentation networks. More details of how DGM is added will be illustrated in the supplementary material. Results in Tab.~\ref{Tab:ade20k} (see \textbf{+DGM}) show that DGM can increase the performance for every backbone model. Admittedly, OCNet and APCNet show better performance on ADE20k. Our model serves a better role in the more challenging unified perceptual parsing where a joint representation for multiple tasks is needed.

\vspace{-2mm}
\paragraph{Top-down message passing.}
To evaluate the role of \textit{TDMP}, we evaluate DGM model without \textit{TDMP} (denoted as \textbf{DGM w/o $\downarrow$}). In the Broden+ dataset, DGM w/o $\downarrow$ shows weaker performance compared with the full model, proving the effectiveness of context modeling of \textit{TDMP}. In the single-task ADE20k, DGM w/o $\downarrow$ performs weaker on DeepLabV3 backbone and achieves even better performance than the full model when UPerNet and HRNetv2 are the backbones. We suspect that the top-down message passing may not provide valuable information to lower-level graph features when only one task is trained and evaluated. In comparison, \textit{TDMP} helps lower-level graph features for better prediction on part segmentation and material segmentation (see Tab.~\ref{Tab:Broden}).

\subsection{Grouping Visualization}
To verify the quality of perceptual grouping, the grouping results are visualized in Fig.~\ref{Fig:group_visualization}. Details of grouping visualization will be illustrated in the supplementary material. As shown in Fig.~\ref{Fig:group_visualization}, DGM gradually merges conceptually-related regions as it goes to higher levels in the hierarchy. For example, in the second row, sofa gradually merges with tables to the main area in the living room.

\subsection{Overhead}
\begin{table}[t]
\centering
\begin{tabular}{c|c|c} \toprule
 Model                 & FLOPs ($\Delta$)    & \#Params ($\Delta$) \\ \midrule
  OCP                  & 161.4G        & 15.179M                 \\
                          RCCA(R=2)                  & 16.5G & 23.931M                    \\
                          \textbf{DGM w/o$\downarrow$}                   & \textbf{9.3G}          & \textbf{3.417M}  \\
                          \textbf{DGM}                    & \textbf{10.8G}          & \textbf{4.468M}  \\ \bottomrule

\end{tabular}
\caption{Compare overhead of contextual modules.}
\label{Tab.overhead}
\vspace{-6mm}
\end{table}

We compare the overhead with other contextual modeling methods: recurrent criss-cross attention (RCCA) module proposed in CCNet~\cite{Huang_2019_ICCV} and object context pooling module (OCP) in OCNet~\cite{Yuan_2018_ArXiv}. For a fair comparison, the size of the input images to all methods is $769 \times 769$. In Tab.~\ref{Tab.overhead}, we show the difference of FLOPs and the number of parameters before and after adding the contextual modeling module to the network. For our proposed DGM model, we use ResNet50~\cite{Xiao_2018_ECCV} as the backbone when evaluating the overhead. The results show that our model has significantly lower overhead compared with non-local base OCP module. Note that RCCA is the state-of-art method targeting at reducing overhead in contextual modeling. Our method beats RCCA module of CCNet. In Tab.~\ref{Tab.overhead}, we also show the overhead of our method without using \textit{TDMP} (DGM w/o $\downarrow$). The result shows that \textit{TDMP} only creates little overhead.

\section{Applications}

We further show that DGM enables novel applications due to the added interpretability of the perceptual grouping process, which is difficult to achieve by using other segmentation networks.

\begin{figure}[t]
  \includegraphics[width=\columnwidth]{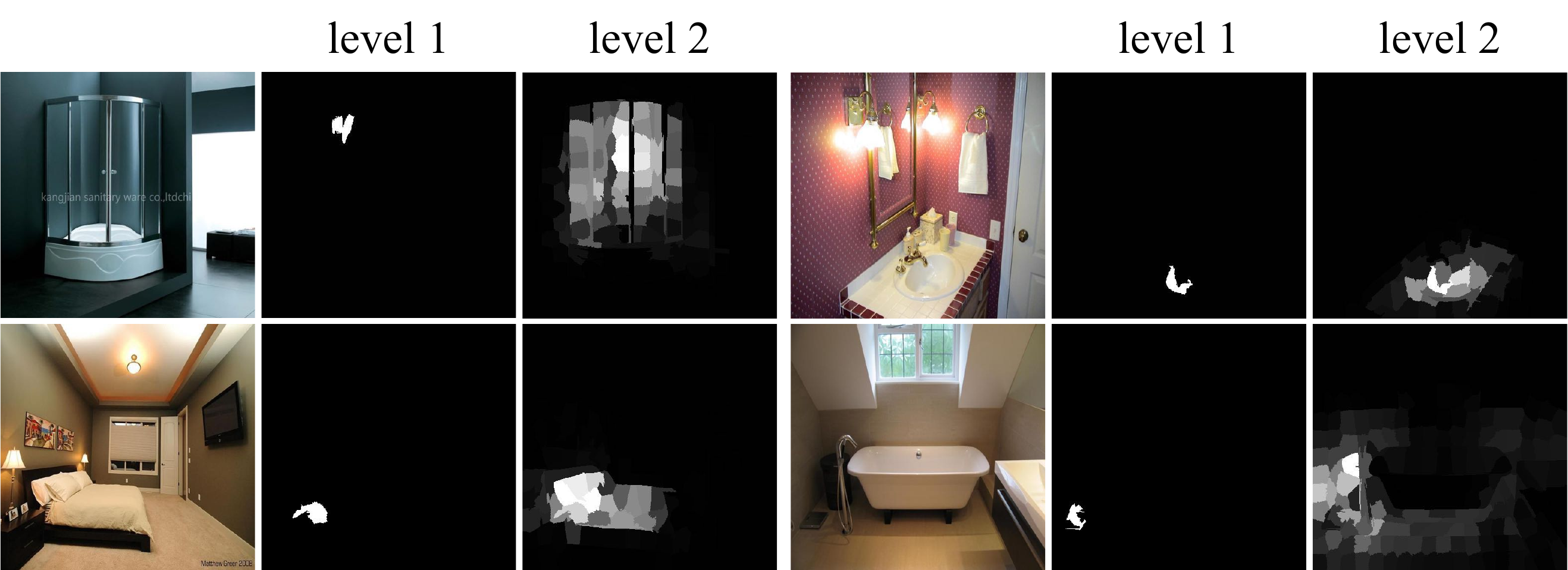}
  \centering
  \caption{Visualization of click propagation. Bottom-right is negative click while others are all positive clicks.}
  \label{Fig:click}
  \vspace{-5mm}
\end{figure}

\begin{figure}[t]
  \includegraphics[width=\columnwidth]{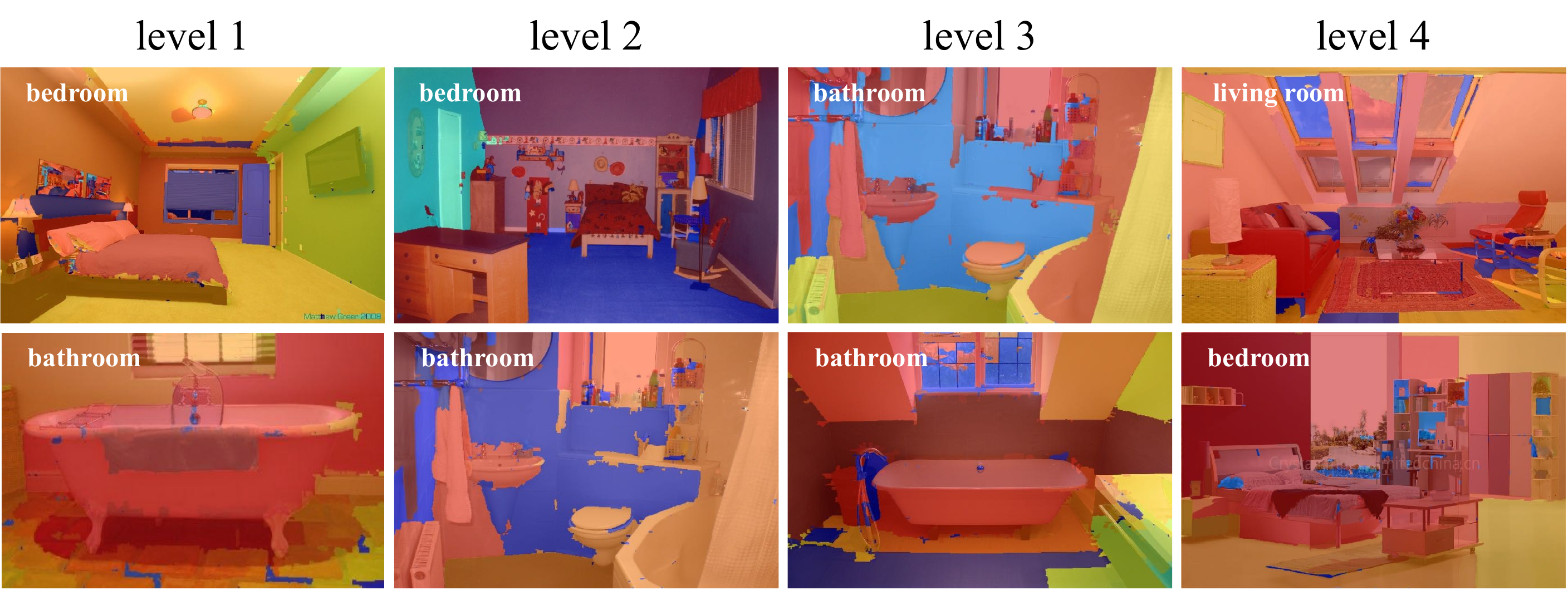}
  \centering
  \caption{Visualization of Grad-CAM. Red-to-blue denotes the decreasing activation. Scene labels are shown.}
  \label{Fig:gradcam}
  \vspace{-4mm}
\end{figure}

\noindent \textbf{Click Propagation.}
In interactive segmentation, a user adds positive click on the object and negative click on the background, which are used to segment the selected instance on the image. One critical process of recent interactive segmentation methods~\cite{Majumder_2019_CVPR, Xu_Ning_2016_CVPR} is augmenting user's click by propagating it to other related areas on the image. Since our model produces a compositional-hierarchical graph, related areas can be dynamically computed through the learning process, rather than treating it as a pre-processing step. As shown in Fig.~\ref{Fig:click}, given a user's click, our model first selects a superpixel. Then, it can propagate to higher levels by using the $\mathbf{P}^l$ defined in Eq.~\ref{Eq.pool_weight}. For example, positive click is propagated to the entire shower kit in Fig.~\ref{Fig:click} top-left, and negative click will not be propagated to the bathtub in Fig.~\ref{Fig:click} bottom-right. More details are in the supplementary material.

\noindent \textbf{Explainability with Grad-CAM.}
We use Grad-CAM on graph~\cite{Pope_2019_CVPR} to localize activated vertices at each level of the hierarchy (more details in supplementary material). By using the gradient back-propagated from the ground-truth scene label, our model localizes semantically discriminative regions on the image. For example, the bed is highlighted with sharp boundary in Fig.~\ref{Fig:gradcam} bedroom.

\section{Conclusion}
\label{Sec.conclusion}
We propose Deep Grouping Model to marry a CNN segmentation network with the perceptual grouping process, which outperforms state-of-the-art methods on unified perceptual parsing task with little overhead. Meanwhile, our proposed model is of good interpretability and is useful in other tasks. We believe such hierarchical graph representation is of great potential to be applied to many other tasks.

\noindent \textbf{Acknowledgments.}
 This work was supported in part by NSF 1741472, 1764415, 1813709, and 1909912. The article solely reflects the opinions and conclusions of its authors but not the funding agents.

{\small
\bibliographystyle{ieee_fullname}
\bibliography{main_bib}
}

\end{document}